\pdfoutput=1
\documentclass[11pt]{article}

\usepackage{EMNLP2022}

\usepackage{times}
\usepackage{latexsym}

\usepackage[T1]{fontenc}
\usepackage[utf8]{inputenc}
\usepackage{microtype}

\usepackage{inconsolata}
\usepackage{multirow}
\usepackage{booktabs}
\usepackage{graphicx}
\usepackage{amsmath}
\usepackage{CJKutf8}

\usepackage[hang,flushmargin]{footmisc}

\DeclareMathOperator{\Sim}{sim}
\DeclareMathOperator{\Enc}{E}

\usepackage{xspace}
\usepackage{xcolor}
\definecolor{amber}{rgb}{1.0, 0.75, 0.0}

\newcommand{\ours}{\textsc{XRICL}\xspace}

\title{\ours: Cross-lingual Retrieval-Augmented In-Context Learning for Cross-lingual Text-to-SQL Semantic Parsing}

\author{Peng Shi$^{\spadesuit}$, Rui Zhang$^{\clubsuit}$, He Bai$^{\spadesuit}$, \and Jimmy Lin$^{\spadesuit}$ \\[1ex]
  $^{\spadesuit}$ University of Waterloo \quad $^{\clubsuit}$ Penn State University \\[1ex]
  \texttt{\{peng.shi,he.bai,jimmylin\}@uwaterloo.ca, rmz5227@psu.edu}
}

\begin{document}
\maketitle
\begin{abstract}
In-context learning using large language models has recently shown surprising results for semantic parsing tasks such as Text-to-SQL translation.
Prompting GPT-3 or Codex using several examples of question-SQL pairs can produce excellent results, comparable to state-of-the-art finetuning-based models.
However, existing work primarily focuses on English datasets, and it is unknown whether large language models can serve as competitive semantic parsers for other languages.
To bridge this gap, our work focuses on cross-lingual Text-to-SQL semantic parsing for translating non-English utterances into SQL queries based on an English schema.
We consider a zero-shot transfer learning setting with the assumption that we do not have any labeled examples in the target language (but have annotated examples in English).
This work introduces the \ours framework, which learns to retrieve relevant English exemplars for a given query to construct prompts.
We also include global translation exemplars for a target language to facilitate the translation process for large language models.
To systematically evaluate our model, we construct two new benchmark datasets, \textsc{XSpider} and \textsc{XKaggle-dbqa}, which include questions in Chinese, Vietnamese, Farsi, and Hindi.
Our experiments show that \ours effectively leverages large pre-trained language models to outperform existing baselines.
Data and code are publicly available at \url{https://github.com/Impavidity/XRICL}.
\end{abstract}

\section{Introduction}

Semantic parsing is the task of translating natural language questions into meaning representations such as Lambda CDS~\cite{liang2013lambda}, Python code~\cite{yin2018structvae}, and SQL~\cite{yu2018spider}.
More recently, Text-to-SQL semantic parsing has attracted attention from academia and industry due to its challenging setup and practical applications.
Cross-lingual Text-to-SQL semantic parsing~\cite{sherborne2021zero,min2019pilot,sherborne2020bootstrapping} aims to translate non-English utterances into SQL queries based on an English schema~(assuming we have an internationalized database), 
enabling users to query databases in non-English languages.
For example, such a system could help people from around the world access the US government's open data\footnote{\url{https://data.gov}} with natural language questions in different languages.

State-of-the-art approaches for Text-to-SQL semantic parsing have been greatly improved by finetuning pre-trained language models as a sequence-to-sequence problem~\cite{scholak2021picard,yin2020tabert,herzig2020tapas,yu2020grappa,yu2021score,shi2020learning}.
More recently, in-context learning with large language models~(LLMs), such as GPT-3~\cite{brown2020language} and Codex~\cite{chen2021evaluating}, has emerged as a new learning paradigm.
This paradigm enables effective few-shot learning without model finetuning, showing its practical and scientific value~\cite{beltagy-etal-2022-zero}.
Recent papers also have shown promising results applying in-context learning to the Text-to-SQL task.
\citet{rajkumar2022evaluating} studied if LLMs are already competitive Text-to-SQL semantic parsers without further finetuning on task-specific training data.
Additionally, \citet{poesia2022synchromesh} and \citet{rubin2021learning} investigated the exemplar retrieval problem for the semantic parsing task.

\begin{figure*}[t!]
    \centering
    \includegraphics[width=0.9\textwidth]{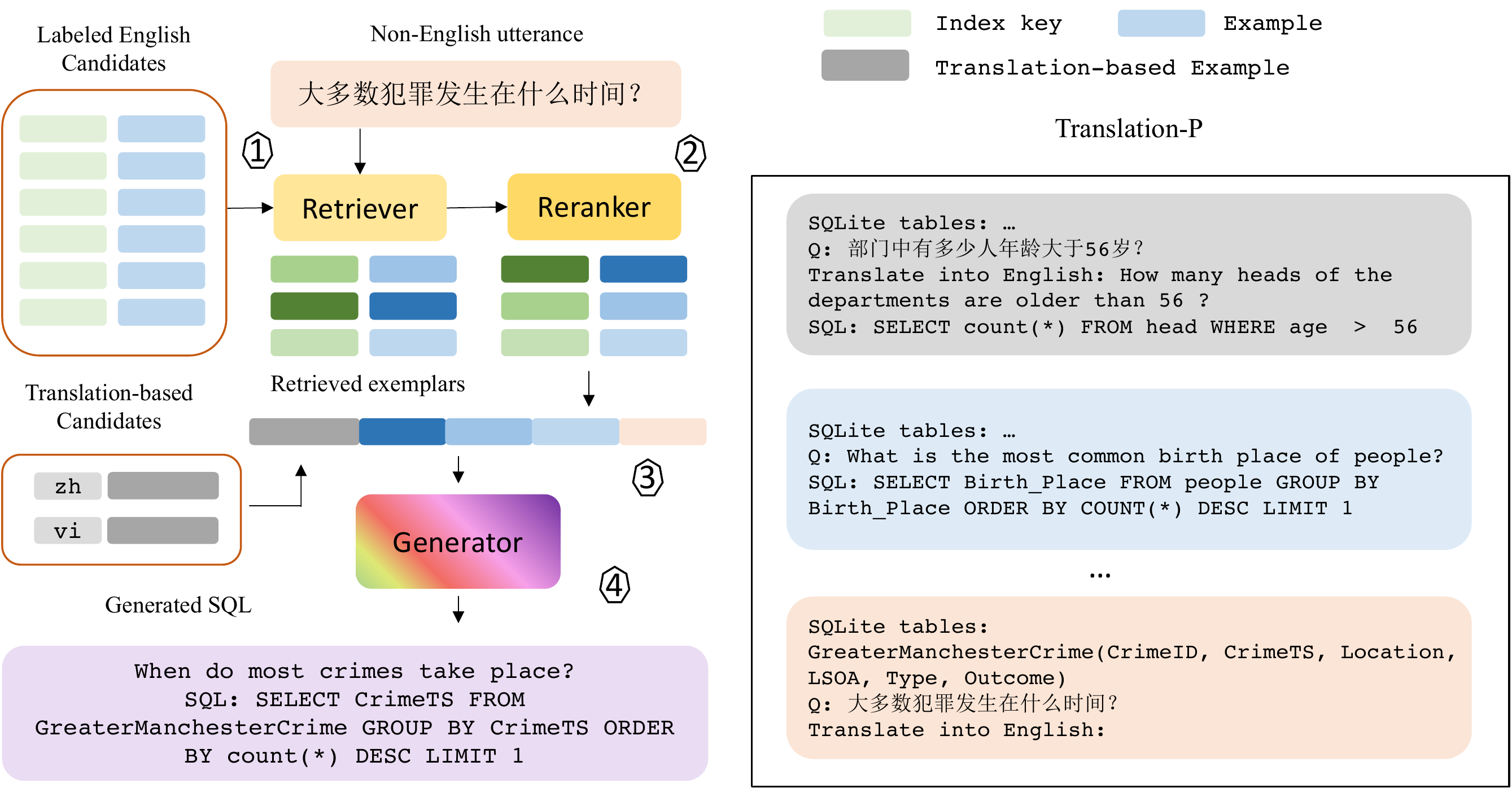}
    \caption{Overview of our proposed \ours framework. Given a labeled English question-SQL candidate pool and the non-English question as input, our framework uses in-context learning with a large pre-trained language model (e.g., Codex) to generate SQL queries in four steps: (1) Cross-lingual Exemplar Retrieval, (2) Exemplar Reranking, (3) Prompt Construction with Translation as Chain-of-Thought, and (4) Inference.}
    \label{fig:arch}
\end{figure*}

However, previous work mostly focused on English utterances, leaving other languages behind.
It is unclear if LLMs are competitive for cross-lingual Text-to-SQL with English exemplars using in-context learning.
Even in the mono-lingual setting (where the exemplars and the query are in the same language), many approaches are not practical beyond English due to the paucity of target language query-SQL exemplars.

To bridge this gap, we propose \ours, a novel framework based on LLMs with in-context learning for cross-lingual Text-to-SQL semantic parsing.
Specifically, the task is to generate SQL queries for non-English queries based on an English schema and an English query-SQL candidate pool. 
Our framework first constructs the context prompt by retrieving the most relevant English query-SQL exemplars for each target language query. 
Since we do not have any training data in the target language, we cannot train a retriever for target queries directly. 
Our solution is to train an English exemplar retriever with mT5~\cite{xue2020mt5} and adopt a model-based cross-lingual transfer method for cross-lingual retrieval.
The English exemplar retriever is trained with feedback from the LLM itself by distilling soft labels~(likelihood).

Our framework introduces an additional exemplar into the LLM's input context, to instruct the model to translate the target query into English and then to translate the English query into SQL; this approach is inspired by recent work on chain-of-thought prompting~\citep{wei2022chain,shi2022language}.
However, in our framework, this additional exemplar is identical for all test queries, which means that we only need a single pair of translations for any English-target language pair, requiring minimal translation effort.

During the inference process, the language model is expected to generate the English translation first and then the SQL query.
In our experiments, we find that our proposed retriever and reranker can improve the LLMs' cross-lingual few-shot in-context learning performance by a large margin, and further improvements can be observed by adding an additional translation exemplar.

We further construct two benchmarks, \textsc{XSpider} and \textsc{XKaggle-dbqa}, to systematically evaluate the proposed framework in many languages.
For \textsc{XSpider}, besides adopting existing work, including \textsc{CSpider}~\cite{min2019pilot} and \textsc{VSpider}~\cite{nguyen2020pilot},
we further translate the \textsc{Spider} dataset into Farsi and Hindi for evaluation.
For \textsc{XKaggle-dbqa}, we translate the English \textsc{Kaggle-dbqa} dataset into Chinese, Farsi, and Hindi.
Experimental results show that our proposed framework improves effectiveness compared to baseline systems.

Our contributions are summarized as follows:
(1) We propose a novel retrieve-rerank framework to improve the exemplar selection process for in-context learning for cross-lingual Text-to-SQL semantic parsing. 
To the best of our knowledge, we are the first to explore the effectiveness of large pre-trained language models for cross-lingual Text-to-SQL semantic parsing. 
(2) We propose to use translation as a chain-of-thought prompt in the inference process, bridging the cross-lingual gap for large language models.
(3) Last, we construct two new benchmarks, \textsc{XSpider} and \textsc{XKaggle-dbqa}, to facilitate evaluation of cross-lingual Text-to-SQL semantic parsing.

\section{Task Formulation}

Given a database where the schema $s$ is in English (denoted as the source language), our task is to translate a non-English (denoted the target language) example $x$ ($x$ includes utterance $u$ and schema $s$)\ into a SQL query $a$.
In this work, we explore large pre-trained language models such as Codex for this Text-to-SQL task with in-context learning.
To support in-context learning, labeled candidates of (utterance, schema, SQL) triples are required.
Since more annotated resources are available in English, we assume that the labeled candidate set $D$ is in English.
Overall, in-context learning is an efficient method to leverage large pre-trained language models without expensive parameter fine-tuning.
Furthermore, the candidate pool can be easily expanded for better generalization to new domains.

\section{The \ours Framework}

Our \ours framework is shown in Figure~\ref{fig:arch}, consisting of four steps:

\smallskip \noindent {(1) \textit{Cross-lingual Exemplar Retrieval}}: Retrieve a list of $N$ English exemplars that are relevant to the input non-English example $x$.

\smallskip \noindent {(2) \textit{Exemplar Reranking}}: Rerank the retrieved $N$ exemplars and use the top $K$ exemplars to construct prompts. 

\smallskip \noindent {(3) \textit{Prompt Construction with Translation as Chain of Thought}}: Construct a prompt consisting of the translation exemplar as a chain of thought, the selected $K$ exemplars, and the input example.

\smallskip \noindent {(4) \textit{Inference}}: Feed the prompt into a pre-trained language model to generate SQL.

\subsection{Cross-lingual Exemplar Retriever}

Given a non-English question, the goal of the cross-lingual exemplar retriever is to find \textit{relevant} exemplars from the English candidate pool efficiently that can improve the predictions of the generators.
Considering that we use labeled examples in English (a high-resource language) as candidates, 
we formulate this step as a cross-lingual retrieval problem, where the test question is in a non-English language.
In this case, traditional term matching methods such as BM25~\cite{robertson2009probabilistic} or BM25 + RM3 query expansion~\cite{Lin_SIGIRForum2018} cannot be applied due to token mismatch.
Instead, we propose to use a bi-encoder for cross-lingual semantic retrieval with model-based zero-shot transfer.
We further improve the retriever with distillation-based training.

\smallskip \noindent \textbf{Model.}
Here, we leverage the popular bi-encoder architecture known as dense passage retriever (DPR)~\cite{karpukhin2020dense},
where the query and candidates are mapped into representation vectors \textit{independently}.
The retriever uses a dense encoder $\Enc_u(\cdot)$ that converts an utterance into a $d$-dimensional vector 
and builds an index over the candidate pool that is used for retrieval.

For a test instance $x$, we use the same dense encoder to map the utterance into a $d$-dimensional vector (denoted the query vector).
Based on the query vector, the closest top $N$ exemplars are retrieved from the pre-built index based on the pre-defined distance function.
Following \citet{karpukhin2020dense}, we define the distance function as 
\begin{equation}
    \Sim(x, z) = \Enc_{u}(x)^{\top}\Enc_{u}(z)
    \label{eq:sim}
\end{equation}
\noindent where $Z$ is the set of candidate exemplars and $z \in Z$.
We use a transformer as the dense encoder, and the average of the contextual embeddings of the utterance tokens is taken as the representation of the encoded text.

\smallskip \noindent \textbf{Model-based Cross-lingual Transfer.}
Considering that we do not have training data in target languages, we adopt a model-based cross-lingual transfer method,
where we leverage the zero-shot cross-lingual transfer ability of multilingual pre-trained transformers such as mBERT~\cite{devlin2018bert}, XLM-Roberta~\cite{conneau2019unsupervised}, mBART~\cite{liu2020multilingual}, and mT5~\cite{xue2020mt5}.
Specifically, we train the dense retriever in the source language,
where both the query utterance and candidate utterances are in English (in our case), 
and apply inference directly on query utterances in the target language 
and retrieve English exemplars in a cross-lingual manner.

\smallskip \noindent \textbf{Distillation-based Training.}
One common practice for bi-encoder training is contrastive learning.
Given a query, positive examples and negative examples are required.
The model is optimized such that examples from the positive class have similar representations and examples from the negative class have different representations.

The key here is how to define positive and negative examples for the semantic parsing task.
Recently, \citet{hu2022context} used the similarity of target meaning representations to first rank the candidates and choose the top-$k$ as positive examples and the bottom-$k$ as negative examples.
Instead of using human-designed relevance metrics, \citet{rubin2021learning} proposed to use a language model to label positive and negative examples for contrastive learning; similar to \citet{hu2022context}, hard labels are used.
Another way to train the bi-encoder is to use a regression-based loss function.
\citet{poesia2022synchromesh} proposed to retrieve exemplars that have relevant program structures (tree edit distance of SQL abstract syntax trees is used as the relevance metric) for the test utterances and the model is optimized with mean-squared error loss for predicting the similarity score.

\begin{figure}[t]
    \centering
    \includegraphics[width=0.4\textwidth]{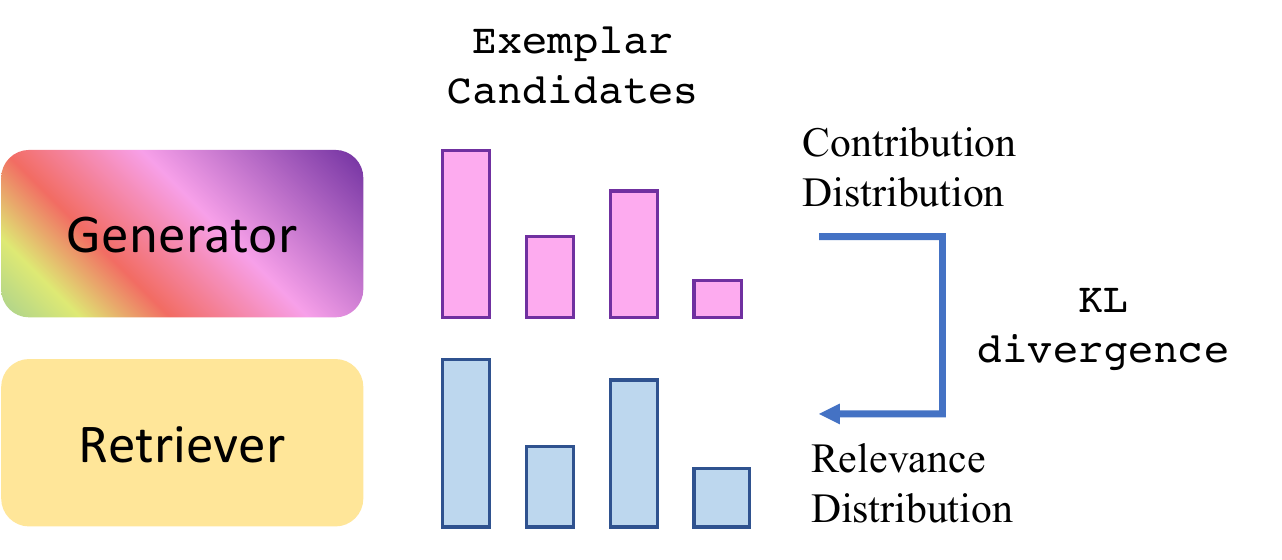}
    \caption{Illustration of distillation-based training. The contribution distribution is the likelihood distribution of the top-$N$ exemplars produced by the LLM. The relevance distribution is the ranking score distribution produced by the retriever.}
    \label{fig:loss}
\end{figure}

As an alternative to these above approaches, we train our retriever by distilling the LLM's scoring function.
This scoring function calculates the ground-truth SQL query's likelihood given an English exemplar $z_k$ and the input utterance $x$, which estimates the importance of this exemplar for parsing the given input utterance.
Hence, we score the retrieved English exemplars with an LLM and optimize the KL divergence between the LLM's ranking scores and the retriever's ranking scores to update the retriever,
as shown in Figure~\ref{fig:loss}.
This retriever is denoted DE-Retriever (Distillation-based Exemplar Retriever).
Intuitively, with the KL divergence loss function, 
the model tries to match the probability of retrieving an exemplar $z_k$ with the contribution of that exemplar to the generated SQL query $a$.

We first obtain $N$ exemplars with the highest scores based on Equation~(\ref{eq:sim}), denoted as $Z_{top-N}$.
Our loss function is defined as:
\begin{equation}
\begin{split}
     \mathcal{L}_{\texttt{distill}} = \texttt{KL}(\:\texttt{SG}(p(z_n \:|\: x, a, Z_{top-N}; G)) \\
    \; \| \; p(z_n\:|\:x, Z; E)),
\end{split}
\end{equation}
\noindent where \texttt{SG} denotes the stop gradient operation, $G$ denotes the generator, and $E$ denotes the retriever encoder.
We further compute $p(z_n\:|\:x, a, Z_{top-N}; G)$ as follows:

\begin{equation}
\begin{split}
    & p(z_n\:|\:x, a, Z_{top-N}) \propto \\
    & p(a\:|\:x, z_n, Z_{top-N};G)\; p(z_n\:|\:x, Z_{top-N})
\end{split}
\end{equation}
\noindent We approximate the posterior under the assumption that we have a uniform prior over the set of retrieved exemplars, so $p(z_n\:|\:x, Z_{top-N})$ is approximated as $\frac{1}{N}$.
We further compute $p(a\:|\:x, z_n, Z_{top-N}; G)$ as:
\begin{equation}
    % p(a\:|\:x, z_n, Z_{top-N};G) = \\
    \frac{\exp(p(a\:|\:x, z_n))}{\sum_{j=1}^{N}{\exp(p(a\:|\:x, z_j))}}
\end{equation}
where $p(a\:|\:x, z_j)$ is computed with the generator.

More specifically, we use example $z_j$ as the prompt and concatenate it with test instance $u$ and target SQL $a$.
Then we feed it to the generator to compute the log probability of each token $\log(p(a_i))$ in the target SQL query $a$;
$p(a\:|\:x, z_j)$ can be computed as $\exp(\sum{\log(p(a_i))})$.

\subsection{Exemplar Reranking}

For tasks such as information retrieval and open-domain question answering, reranking is widely adopted to further improve retrieval results by incorporating a reranker.
Such a two-stage procedure is also useful in a variety of natural language processing tasks.
In this work, following the retrieve-and-rerank idea, we propose to incorporate an exemplar reranker in our framework.
This reranker can leverage token-level interactions between the utterances to better rank the exemplars.

More specifically, the query utterance $u$ and the candidate utterance $u_z$ are concatenated together with special tokens: \texttt{[CLS] $u$ [SEP] $u_z$ [SEP]}.
The tokenized input is fed into a transformer model.
An MLP with sigmoid activation is applied on top of the contextual embedding of the \texttt{[CLS]} token to obtain the relevance score of the candidate example~\cite{Lin_etal_2021_ptr4tr}.
Sigmoid cross-entropy loss is used and the model is optimized to produce a relevance score as $p(a|x,z_n, Z_{top-N}; G)$.
This reranker is denoted DE-Reranker~(Distillation-based Exemplar Reranker).

\subsection{Prompt Construction with Translation as Chain of Thought}

From the input instance $x$ and the list of retrieved-and-reranked exemplars $Z$, we construct the augmented query by concatenating exemplars with the input instance following previous work~\cite{hu2022context,rubin2021learning,poesia2022synchromesh,liu2021makes,brown2020language,pasupat2021controllable}.
For the exemplar, we linearize the table schema, the question, and the SQL query.
The exemplars are sorted by relevance score in descending order.
For the test instance, only the table schema and the question are linearized.
We denote this prompting approach Vanilla-P.

\smallskip \noindent \textbf{Translation as Chain of Thought}:
Recent work on chain-of-thought prompting is designed to solve the multi-step reasoning problem by providing intermediate reasoning steps before the final answer in the prompt~\cite{wei2022chain}.
Inspired by this, we use the translation pair (from non-English to English in our case) as an intermediate step for cross-lingual semantic parsing inference.

Specifically, a translation-based exemplar is inserted in front of $Z$.
For example, in the right part of Figure~\ref{fig:arch}, the grey box contains the Chinese version of the translation as a chain-of-thought prompt.
The question in the prompt is in the target language, followed by an instruction \texttt{Translate into English} and the English translation of the question. 
Note that this translation-based exemplar is shared among all the test instances in that language, as shown in the left part of Figure~\ref{fig:arch}.
The translation-based examples are indexed by the language code, such as \texttt{zh} and \texttt{vi}.
In this way, it only requires minimal translation effort to build the global translation-based exemplar.
We denote this prompting approach Translation-P.

\subsection{Inference}

For inference, we feed the constructed prompt to a large pre-trained language model to generate the target SQL query with greedy decoding.
In this work, we consider \textbf{Codex}~(Codex-Davinci-001)~\cite{chen2021evaluating} because it has shown superior performance for the English Text-to-SQL task~\cite{poesia2022synchromesh}.

\begin{table*}[t]
    \centering
    \small
    \begin{tabular}{lccccccccc}
    \toprule
    \multirow{2}{*}{\textbf{Model}}  & \multicolumn{1}{c}{\textbf{zh-full}} &
    \multicolumn{2}{c}{\textbf{zh}} & \multicolumn{2}{c}{\textbf{vi}} & \multicolumn{2}{c}{\textbf{fa}} & \multicolumn{2}{c}{\textbf{hi}} \\ \cmidrule(l){2-2}\cmidrule(l){3-4}\cmidrule(l){5-6}\cmidrule(l){7-8}\cmidrule(l){9-10}
    & EM & EM & TS & EM & TS & EM & TS & EM & TS  \\
    \toprule
    (1)~mT5 zero-shot & 39.7 & 47.9 & 48.4 & 42.1 & 40.1 & 41.3 & 39.5 & \textbf{41.2} & \textbf{39.7} \\
       (2) mUSE & 38.4 & 43.0 & 46.8 & 31.8 & 33.4 & 28.9 & 31.1 & 22.2 & 23.7 \\
       (3) mSBERT & 37.9 & 41.3 & 47.1 & 34.6 & 33.5 & 29.3 & 31.8 & 22.0 & 22.3 \\
       \midrule
       (4) mT5-encoder & 44.4 & 48.1 & 51.4 & 41.3 & 39.5 & 38.4 & 38.5 & 28.6 & 27.0 \\
       (5) DE-Retriever & 46.0 & 50.4 & 53.9 & 42.2 & 40.7 & 38.2 & 40.0 & 29.9 & 27.9 \\
       (6) DE-R$^2$ & 46.4 & 52.1 & 55.3 & \textbf{44.4} & 41.9 & 40.0 & 40.6 & 30.0 & 28.2 \\
       (7) + Translation-P & \textbf{47.4}& \textbf{52.7} & \textbf{55.7} & 43.7 & \textbf{43.6} & \textbf{43.2} & \textbf{45.1} & \textbf{32.6} & \textbf{32.4} \\
      \bottomrule
    \end{tabular}
    \vspace{1.5mm}
    \caption{Results on the \textsc{XSpider} dev set. ``zh-full'' and ``zh'' are two different splits from \textsc{CSpider}~\cite{min2019pilot}. EM and TS are exact match accuracy and test suite accuracy, respectively.
    Entry (5) is based on the DE-Retriever with Vanilla-P.
    Entry (6) is based on the DE-Retriever and DE-Reranker (denoted as DE-R$^2$) with Vanilla-P.
    Entry (7) is based on DE-R$^2$ with Translation-P.}
    \vspace{1.5mm}
    \label{tab:main_xspider}
    
\end{table*}

\section{Experimental Settings}
In this section, we describe the datasets, implementation details, and baselines for our experiments.

\subsection{Datasets}

We create two benchmarks, \textsc{XSpider} and \textsc{XKaggle-dbqa}, by translating existing English Text-to-SQL datasets into other languages and evaluate our methods on these two benchmarks.

\smallskip \noindent \textbf{\textsc{XSpider}}: \textsc{CSpider}~\cite{min2019pilot} and \textsc{VSpider}~\cite{nguyen2020pilot} are \textbf{C}hinese~(zh) and \textbf{V}ietnamese~(vi) cross-domain Text-to-SQL datasets translated from \textsc{Spider}~\cite{yu2018spider}.
More specifically, we use the English \textsc{Spider} training set as the candidate pool and training data for retriever-reranker models.
We use the development sets of \textsc{CSpider} and \textsc{VSpider} for cross-lingual evaluation.
We further translate the \textsc{Spider} development set into Farsi~(fa) and Hindi~(hi) for a more comprehensive evaluation.

\smallskip \noindent \textbf{\textsc{XKaggle-dbqa}}:
This is a recently constructed dataset for more realistic and challenging Text-to-SQL evaluation.
The dataset is based on 8 databases from Kaggle.
We translate the questions into Chinese~(zh), Farsi~(fa), and Hindi~(hi) for cross-lingual evaluation.
We use the English \textsc{Spider} training set as the candidate pool. 

\subsection{Experimental Details}

For the exemplar retriever, we use 24-layer transformers
initialized with the parameters of the mT5 encoder that is then fine-tuned on the English \textsc{Spider} dataset for the Text-to-SQL task.
For the exemplar reranker, we use InfoXLM~\cite{chi2020infoxlm} as the starting point.
We train the retriever and reranker on the English \textsc{Spider} dataset and then apply both models to cross-lingual retrieval and reranking in a zero-shot fashion. 
For the Codex configuration, we use greedy decoding by setting the temperature to zero.
We use $N=16$ and $K=8$ for all experiments, which means that the DE-Retriever first retrieves 16 exemplars from the candidate pool and the DE-Reranker produces the top 8 exemplars for prompt construction.

In terms of evaluation metrics, we use \textbf{E}xact \textbf{M}atch~(EM) accuracy for both the \textsc{XSpider} benchmark and the \textsc{XKaggle-dbqa} benchmark.
Following \citet{zhong2020semantic}, we report the \textbf{T}est-\textbf{s}uite~(TS) accuracy.
Only the datasets that are aligned with the \textsc{Spider} dev set can be evaluated with TS accuracy, so the \textsc{XKaggle-dbqa} benchmark is not applicable.
Because the \textsc{CSpider} dev set is only partially aligned to the \textsc{Spider} dev set, the full \textsc{CSpider} (zh-full) dev set can be only evaluated with EM accuracy. 
We collect a subset of the \textsc{CSpider} dev set (zh) whose queries are aligned with the English \textsc{Spider} dev set, and further evaluate these using TS accuracy.

\subsection{Baselines}

\noindent \textbf{mT5 zero-shot transfer} is a baseline model that is trained with the English \textsc{Spider} training set.
The model is based on the pre-trained sequence-to-sequence multilingual language model mT5-large~\cite{xue2020mt5}.
This model has zero-shot cross-lingual transfer ability, with which the model can directly handle non-English utterances.

\smallskip \noindent \textbf{mUSE and mSBERT} are baselines that use unsupervised retrievers to obtain exemplars:\ multilingual Universal Sentence Encoder~\cite{yang2019multilingual} and multilingual Sentence-BERT~\cite{reimers-2019-sentence-bert}.
Prompts are then constructed for in-context learning with Codex.

\section{Results}

\subsection{Results on \textsc{XSpider}}

Results on \textsc{XSpider} are shown in Table~\ref{tab:main_xspider}.
We report the EM and TS accuracy.
For the full \textsc{CSpider} dataset (zh-full), since TS Accuracy is not supported, we only report EM accuracy.
We report both TS and EM accuracy on the subset of \textsc{CSpider}.
Entry (1) reports the zero-shot performance of the mT5 model that is trained on the English \textsc{Spider} dataset.
On \texttt{zh-full}, \texttt{vi}, \texttt{fa}, and \texttt{hi}, the mT5 zero-shot method obtains on average 41.1 EM accuracy and 39.8 TS accuracy (average TS accuracy is computed without \texttt{zh-full} because the metric cannot be computed on the full \textsc{CSpider}).

From entry (2) to entry (7), the methods are based on in-context few-shot learning. 
For entries (2--6), the prompting method is Vanilla-P.
For entry (7), prompting with Translation-P is applied.

\vspace{1.5mm}
With unsupervised exemplar retrievers such as mUSE and mSBERT, shown in entries (2) and (3), Codex performs worse than mT5 zero-shot transfer, especially for Farsi (39.5$\rightarrow$31.1/31.8 on TS accuracy) and Hindi (39.7$\rightarrow$23.7/22.3 on TS accuracy).
By switching the unsupervised exemplar retriever to the mT5-encoder, which is the encoder component of the fine-tuned mT5 model, the effectiveness of Codex improves by a large margin.
For example, on the \textsc{CSpider} subset, TS accuracy improves to 51.4 from 47.1, outperforming mT5 zero-shot performance by 3 points.
This indicates that the exemplar retrieval component is essential to take advantage of the competitive performance of LLMs such as Codex.
For languages such as Vietnamese and Farsi, Codex is comparable to mT5 zero-shot transfer, while for Hindi, there is still a large gap (39.7 vs.\ 27.0 on TS accuracy).

By applying our proposed distillation based retriever-reranker pipeline~(denoted as DE-R$^2$) for retrieving exemplars, impressive improvements can be observed in all four languages by comparing entry (6) with entry (4).
Our end-to-end results are shown in entry (7), where we see that our proposed framework achieves the best results for most of the languages (except Vietnamese EM accuracy) in the in-context learning setting.

Comparing the best results of in-context learning with mT5 zero-shot results, we can see that Codex can achieve better performance in Chinese, Vietnamese, and Farsi.
For example, \ours outperforms mT5 zero-shot by 7.7 EM accuracy on the full dev set of \textsc{CSpider}.
One exception is Hindi, where the best in-context learning performance cannot match mT5 zero-shot transfer.
One possible explanation is that Codex has weaker modeling ability in Hindi because less Hindi data were accessible during the training.

\begin{table}[t]
    \centering
    \small
    \begin{tabular}{lccc}
    \toprule
    \textbf{Model} & \textbf{zh} & \textbf{fa} & \textbf{hi} \\
    \midrule
    (1)~mT5 zero-shot & 9.7 & 8.1 & 7.6 \\
    (2)~mUSE & 20.7 & 12.4 & 16.2 \\
    (3)~mSBERT & 14.7 & 13.0 & 11.9 \\
    \midrule
    (4) mT5-Encoder & 22.2 & 16.8 & 16.2 \\
    (5) DE-Retriever & 26.5 & 18.4 & 16.8 \\
    (6) DE-R$^2$ & 27.0 & 18.4 & 17.8 \\
    (7) + Translation-P & \textbf{28.1} & \textbf{20.0} & \textbf{19.5} \\
    \bottomrule
    \end{tabular}
    \caption{Results on the \textsc{XKaggle-dbqa} test set. We report exact match (EM) accuracy.}
    \label{tab:xkaggledbqa_main_table}
\end{table}

\subsection{Results on \textsc{XKaggle-dbqa}}

There is agreement by researchers today that \textsc{XKaggle-dbqa} is a more realistic evaluation for the Text-to-SQL parsing task.
The databases are real-world databases with abbreviated column names.
We use the training set of English \textsc{Spider} as the candidate pool.
In this case, both the model's generalization ability and its cross-lingual transfer capability can be tested.

The \textsc{XKaggle-dbqa} results are shown in Table~\ref{tab:xkaggledbqa_main_table}.
Entry (1) shows the zero-shot cross-lingual cross-domain transfer performance of the mT5 model trained on the English \textsc{Spider} dataset.
For example, on Chinese \textsc{Kaggle-dbqa}, mT5 only obtains 9.7 EM accuracy.
For comparison, mT5 reach 20.0 EM accuracy on the English test set in a zero-shot fashion, outperforming the previous state of the art obtained by RAT-SQL~\cite{wang-etal-2020-rat} with 18.4 EM accuracy~\cite{lee2021kaggledbqa} using column descriptions and model adaptation.
This indicates that the mT5 model is more robust than RAT-SQL on domain transfer.
However, the effectiveness degrades drastically when mT5 is applied to non-English languages.
The mT5 zero-shot method on average obtains only 8.5 EM accuracy in the three languages.

For the Codex-based in-context learning methods, the results are shown in entries (2--7).
With unsupervised retrieval methods such as mUSE,  Codex can reach 20.7 EM accuracy in Chinese, improving over the zero-shot mT5 baseline.
Comparing entries (2) and (3), there is no clear winner for these two unsupervised retrieval methods.
Our end-to-end results are shown in entry (7), which achieves state-of-the-art performance on the \textsc{XKaggle-dbqa} benchmark, with 22.5 EM accuracy on average, which is better than the mT5 zero-shot method.
For example, on Chinese \textsc{Kaggle-dbqa}, our framework obtains an 18.4 point improvement over mT5 zero-shot transfer.

\section{Analysis}
\label{analysis}

\subsection{Effectiveness on English Text-to-SQL}

We show that our model is comparable to other in-context learning methods for English semantic parsing.
Through this comparison, we show that our framework is built on a competitive backbone for Text-to-SQL.
We use the DE-Retriever as the backbone model in the ablation study and compare with three recent methods, described as follows:
\citet{rubin2021learning} used hard labels obtained from the generator to train the retriever.
\citet{poesia2022synchromesh} used the tree edit distance of SQL queries as a similarity function:\ a smaller distance means better exemplar quality for the specific test instance.
The ranking model is optimized to predict the target SQL pair tree edit distance based on the utterance pair. 
\citet{rajkumar2022evaluating} designed an efficient prompt that leverages table contents for zero-shot Text-to-SQL.
We refer the reader to the original papers for more details.

Table~\ref{tab:spider_table} shows the results on the \textsc{Spider} development set.
Our backbone system (DE-Retriever + Codex Generator) obtains 53.5 EM accuracy and 60.3 EX accuracy, which is comparable to the 60.0 EX accuracy reported by \citet{poesia2022synchromesh}.
Comparing to \citet{rajkumar2022evaluating}, our system obtains comparable TS accuracy~(56.3 vs.\ 55.1).

\begin{table}[t!]
    \centering
    \small
    \begin{tabular}{lccc}
    \toprule
    \textbf{Model} & \textbf{EM} & \textbf{EX} & \textbf{TS} \\
    \midrule
    \citet{rubin2021learning}~(our impl.) & 48.5 & 53.5 & 50.3 \\
    \citet{poesia2022synchromesh}  & - & 60.0 & - \\
    \citet{rajkumar2022evaluating} & - & \textbf{67.0} & 55.1 \\
    DE-Retriever (Ours) & 53.5 & 60.3 & \textbf{56.3} \\
    \bottomrule
    \end{tabular}
    \caption{Results on the English \textsc{Spider} development set. Our system achieves results comparable to other state-of-the-art in-context learning methods for English Text-to-SQL. EM:\ Exact Match Accuracy. EX:\ Execution Accuracy. TS:\ Test-suite Accuracy~\cite{zhong2020semantic}.}
    \label{tab:spider_table}
\end{table}

\subsection{Effectiveness of DE-R$^2$}

We analyze the effectiveness of DE-R$^2$ on the \textsc{XSpider} benchmark and the \textsc{XKaggle-dbqa} benchmark.
By comparing entries (5) and (4) in Table~\ref{tab:main_xspider} and Table~\ref{tab:xkaggledbqa_main_table},
we can observe that the DE-Retriever can improve over the mT5-encoder baseline in most of the languages (except EM accuracy in Farsi).
Comparing entries (6) and (5), we find that the reranker can further improve the EM accuracy and the TS accuracy.
This indicates that our \ours framework is effective in selecting good exemplars as prompts.

\subsection{Effectiveness of Chain-of-Thought Prompt}

By comparing entries (7) and (6) in Table~\ref{tab:main_xspider} and Table~\ref{tab:xkaggledbqa_main_table},
we find that Translation-P can further improve the semantic parsing ability of Codex on top of DE-R$^2$, except EM accuracy for Vietnamese.

\subsection{Oracle Performance}

It is interesting to investigate the upper bound of Codex on cross-lingual Text-to-SQL semantic parsing.
We design two pipelines to experiment with the capabilities of Codex when an oracle is available (i.e., the target SQL query is accessible to help the retrieval and reranking).
We experiment with two different oracles:

\smallskip \noindent \textbf{Template Oracle}:
We retrieve exemplars using the \textit{gold} parse.
The template is extracted from the target SQL query and only exemplars with the same SQL template are retrieved.
This is based on the assumption that utterances with the same SQL templates share the same query intent and the generator can benefit from these exemplars.

\smallskip \noindent \textbf{Template Oracle + Codex LM oracle}:
Here we introduce an oracle from the generator~(Codex) into the pipeline.
More specifically, we replicate the training process in the testing phase.
The exemplars with the same SQL templates are first retrieved.
For each retrieved exemplar, we use Codex to compute its contribution to the test instance as the reranking score.
We then use the top-$k$ as the exemplars.

\begin{table}[t!]
    \centering
    \small
    \begin{tabular}{lccc}
    \toprule
    \multirow{2}{*}{\textbf{Model}}  & \multicolumn{1}{c}{\textbf{zh-full}} &
    \multicolumn{2}{c}{\textbf{zh}} \\ \cmidrule(l){2-2}\cmidrule(l){3-4}
    & EM & EM & TS   \\
    \toprule
       (1) DE-R$^2$ + Translation-P & 47.4 & 52.7 & 55.7  \\
       (2) T-Oracle & 46.3 & 52.6 & 57.6 \\
       (3) TG-Oracle & 52.5 & 58.0 & 62.2  \\
      \bottomrule
    \end{tabular}
    \caption{Results with oracles:\ T-Oracle is the Template Oracle and TG-Oracle is the Template+Generator Oracle. EM accuracy and TS accuracy are reported.}
    \label{tab:cspider_oracle}
\end{table}

\medskip \noindent
The experimental results are shown in Table~\ref{tab:cspider_oracle}.
Comparing entries (1) and (2), we can observe that our \ours framework can outperform the Template Oracle in terms of EM accuracy on the full dataset and is comparable on the subset.
Template Oracle + Codex LM Oracle reaches 52.5 on the full dataset and 58.0 on the subset in terms of EM accuracy.
This suggests that signals from the Codex LM are useful and that there is additional room for improvement in our framework.

\section{Related Work}

\noindent \textbf{In-context Learning}:
In-context learning is a relatively new paradigm for zero-shot and few-shot learning with large-scale pre-trained language models, first proposed in GPT-3~\cite{brown2020language}.
In-context learning for semantic parsing has been intensively investigated recently~\cite{pasupat2021controllable,rubin2021learning,shin2021few,rajkumar2022evaluating,hu2022context,xie2022unifiedskg,chen2021evaluating,poesia2022synchromesh}.
However, most of the work considers only English, without examining the cross-lingual ability of the proposed methods.
\citet{winata2021language} evaluated the multilinguality of pre-trained language models on non-English multi-class classification with in-context learning.
However, their task is simpler than semantic parsing tasks such as ours.
To the best of our knowledge, we are the first to explore cross-lingual Text-to-SQL semantic parsing under the in-context learning setting.

\smallskip \noindent \textbf{Cross-lingual Semantic Parsing}:
Cross-lingual semantic parsing aims to handle user utterances from multiple languages and translate them into formal representations.
Recent advances can be categorized into two threads:\ multilingual dataset creation and model development.

For example, \citet{bai-etal-2018-source} adapted a Chinese dialogue parsing dataset into English.
\citet{min2019pilot} and \citet{nguyen2020pilot} adapted the English Text-to-SQL dataset \textsc{Spider}~\cite{yu2018spider} into Chinese and Vietnamese, which are used in this work for evaluation.
Some multilingual datasets with different formal representations have also been created, such as SPARQL~\cite{cui2021multilingual} and TOP~\cite{li2020mtop}.

In terms of model development, \citet{shao2020multi} is the most relevant to our work, which leveraged bilingual input for the semantic parsing task.
However, they used RNN models and focused on multilingual representation alignment with pre-training.
Instead, our work focuses on representation mixup with large multilingual pre-trained models.
Improving cross-lingual zero-shot transfer is another direction~\cite{sherborne2020bootstrapping,sherborne2021zero,sherborne2022meta}.

\smallskip \noindent \textbf{Multilingual and Cross-lingual Retrieval}:
In multilingual retrieval, the task is to retrieve relevant documents where the user queries and the corpora are in the same language.
Recent work takes advantage of cross-language transfer using pre-trained multilingual models~\cite{Shi_etal_FindingsEMNLP2020,shi2021cross,zhang2022towards,zhang2021mr}.
For example, \citet{shi2021cross} used DPR to retrieve documents based on ad-hoc queries in six languages.
On the other hand, cross-lingual retrievers help users find relevant documents in languages that are different from that of the queries.
This task has a long history that goes back several decades~\cite{Nie_2010}, but recent work includes~\citet{zhang2021mind,litschko2022cross,sun2020clirmatrix}.
For instance, \citet{asai2020xor} created a cross-lingual open-domain question answering dataset where the system is required to retrieve passages from different languages to answer user questions.

\section{Conclusion}

In this work, we proposed the \ours framework that improves in-context learning for cross-lingual Text-to-SQL semantic parsing.
The retrieve-and-rerank models that we propose can learn signals from large pre-trained models (Codex) to improve the quality of selected exemplars, which can further benefit the generator.
By integrating prompts inspired by chain of thought, our proposed Translation-P method can bridge the cross-lingual gap for the generator.
Extensive experiments on \textsc{XSpider} and \textsc{XKaggle-dbqa} demonstrate the effectiveness of our framework, which obtains state-of-the-art performance on few-shot in-context learning in most of the datasets, thus unlocking the potential of Codex.

\section{Limitations}

Our work is based on the large language model Codex, which is not open-sourced.
To replicate our experiments, an application to OpenAI for Codex API access is required.
Due to annotation costs, we were unable to evaluate on more languages than those described in this paper.
In the future, we plan to collect more data to investigate Codex performance on different language families.

\section*{Acknowledgements}

This research was supported in part by the Natural Sciences and Engineering Research Council (NSERC) of Canada, Compute Ontario, and Compute Canada.

\bibliography{custom}
\bibliographystyle{acl_natbib}

\end{document}